\documentclass[11pt]{article}

\usepackage[preprint]{acl}

\usepackage{times}
\usepackage{latexsym}

\usepackage[T1]{fontenc}

\usepackage[utf8]{inputenc}

\usepackage{microtype}

\usepackage{inconsolata}

\usepackage{graphicx}

\usepackage{amsmath}
\usepackage{amssymb}
\usepackage{makecell}
\usepackage{booktabs}
\usepackage{listings}
\usepackage{xcolor} 
\usepackage{multirow}   
\usepackage{array}      

\newcommand{\vpara}[1]{\vspace{1.5ex}\noindent\textbf{#1}}

\title{Optimizing Native Sparse Attention with Latent Attention and Local Global Alternating Strategies}

\author{
    Yuxuan Hu\textsuperscript{\rm 1,\rm 2},\ Jianchao Tan\textsuperscript{\rm 2},\ Jiaqi Zhang\textsuperscript{\rm 2},\ Wen Zan\textsuperscript{\rm 2}, Pingwei Sun\textsuperscript{\rm 2} \\
    \  {\bf Yifan Lu}\textsuperscript{\rm 2},\ {\bf Yerui Sun}\textsuperscript{\rm 2},\ {\bf Yuchen Xie}\textsuperscript{\rm 2},\ {\bf Xunliang Cai}\textsuperscript{\rm 2},\ {\bf Jing Zhang}\textsuperscript{\rm 1}\thanks{\ \ Corresponding author.} \\
    \textsuperscript{\rm 1}School of Information, Renmin University of China, Beijing, China \\
    \textsuperscript{\rm 2}Meituan, Beijing, China \\
    \texttt{\{huyuxuan1999,zhang-jing\}@ruc.edu.cn}
}

\begin{document}
\maketitle
\begin{abstract}
In this work, we conduct a systematic analysis of Native Sparse Attention (NSA) and propose targeted improvements that enhance long-context modeling. A key insight is that alternating between local (sliding-window) and global (compression/selective) attention across layers, rather than using fixed patterns, enables more effective propagation of long-range dependencies and substantially boosts performance on long-sequence tasks. Meanwhile, we further refine NSA’s branches with Latent Attention that the sliding-window branch is enhanced with Multi-head Latent Attention (MLA) while compression and selective branches adopt Group-head Latent Attention (GLA). These changes reduce KV-cache memory by 50\% versus NSA while improving the model's common-sense reasoning and long-text understanding capabilities. Experiments on models from 340M to 1.3B parameters (trained on 15B and 100B tokens) show our method matches or exceeds full attention and native sparse attention in both common-sense reasoning and long-context understanding tasks.
\end{abstract}

\section{Introduction}

Benefiting from the application of transformer-based large language models~\cite{deepseek-v3, longcat-flash} (LLMs) in deep reasoning~\cite{deepseek-r1, longcat-flash-thinking}, multi-turn agent systems, and codebase-level code comprehension, the research community has been increasingly focusing on the capabilities of LLMs in handling long-context inputs and test-time computation. As sequence lengths increase, the high computational complexity of the attention module has emerged as a significant efficiency bottleneck, constraining the further development of LLMs and necessitating the design of more efficient model architectures. 

Given the inherent sparsity of softmax attention~\cite{sparse-attention}, sparse attention has emerged as a promising strategy to accelerate attention computations, especially under long-context scenarios. In such cases, only a subset of critical key-value pairs interacts with the query at each generation step. Existing research has proposed a variety of training-free and post-training methods for selecting these key-value pairs, including KV-cache eviction techniques~\cite{stramingllm}, as well as index-based~\cite{quest-tang2024, seerattention-gao2025}, sampling-based~\cite{magicpig-chen}, clustering-based~\cite{clusterkv-liu2025}, and hash-based approaches~\cite{hashattention-desai} for KV-cache selection. Despite their significant potential, these training-free and post-training techniques fail to fully explore model sparsity. To this end, recent research has introduced natively trainable sparse attention mechanisms~\cite{moba-lu2025, nsa-yuan2025}, among which native sparse attention~\cite{nsa-yuan2025} (NSA) stands out as the most widely recognized and promising approach. 

\begin{figure}[t]
    \centering
    \includegraphics[width=\linewidth]{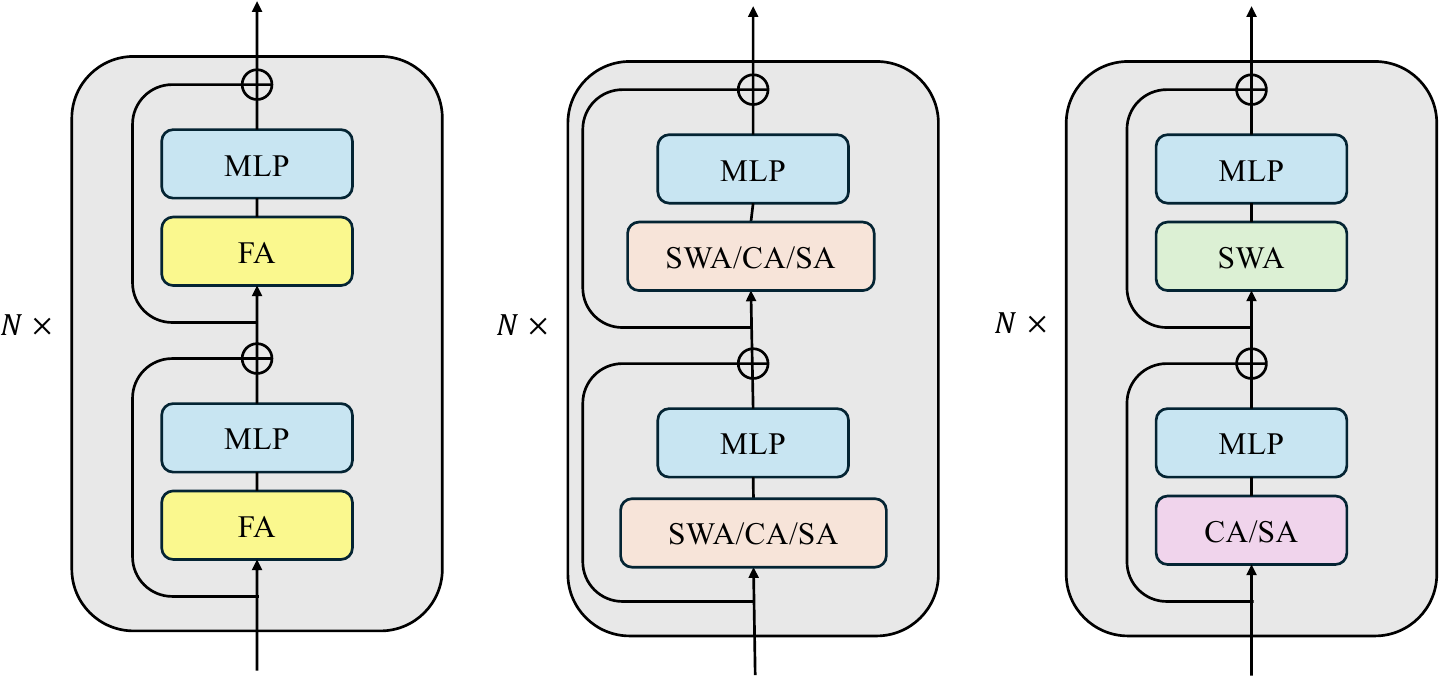}
    \caption{Overview of Llama-like Models Incorporating Full Attention, Native Sparse Attention, and Alternating Sparse Attention.\label{fig: gqa-nsa-ASA}}
    \label{fig: ASA}
\end{figure}

\begin{figure*}[t]
    \centering
    \includegraphics[width=\textwidth]{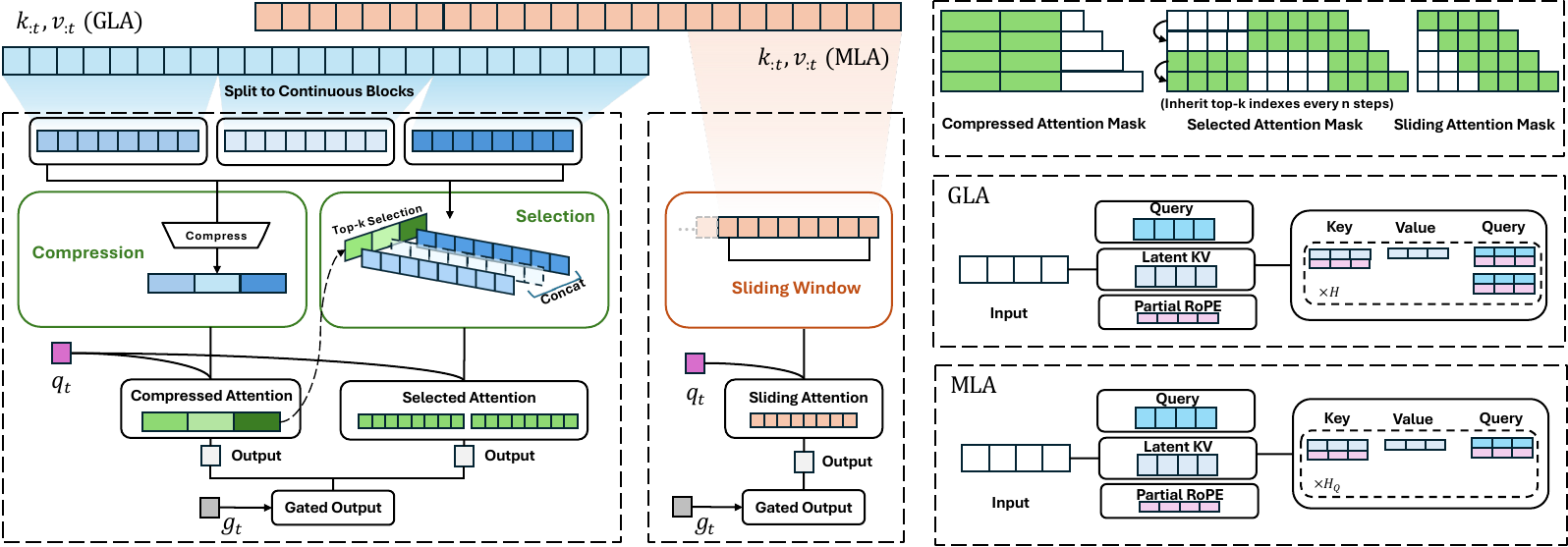}
    \caption{Overview of ASA’s architecture. In ASA, consecutive attention layers alternate between compressed selective attention and sliding window attention. Furthermore, GLA and MLA replace the GQA mechanism used in NSA to enhance model expressiveness. To improve training efficiency, every fourth token in the compressed selective attention module inherits the block index of the first token in its block.}
    \label{fig: ASA}
\end{figure*}

In the NSA framework, dense attention is decomposed into three components: sliding-window attention, compressed attention, and selective attention. Through experimental analysis of these branches, we observe that sliding-window attention plays the dominant role in common sense reasoning, while compression and selective attention primarily serve to enrich the model with global contextual information. Additionally, compared to using the same sparsity level across all layers, we found that an imbalanced sparsity distribution yields better performance on long retrieval tasks.

Building on this insight, we improve NSA by introducing targeted enhancements tailored to the distinct functional roles of each attention branch. The outcome is our proposed method, Alternating Sparse Attention (ASA), a sparse attention architecture explicitly designed for efficient and effective language modeling. As shown in Figure~\ref{fig: gqa-nsa-ASA}, ASA structures attention layers at the individual layer level into two complementary types: sliding-window attention, which effectively models local contextual information, and compress/selective attention, which efficiently captures long-range global context. These two mechanisms are alternated in a strict one-to-one pattern across layers, ensuring a balanced and synergistic representation of both local and global information throughout the model.

On the other hand, NSA is originally implemented based on grouped query attention (GQA). While GQA demonstrates strong effectiveness, certain attention mechanisms such as multi-head latent attention (MLA) have shown even better performance. To further enhance NSA, we replaced GQA with MLA in ASA. However, MLA is equivalent to multi head attention (MHA) during training, which presents challenges when integrating MLA with sparse attention mechanisms. To address this issue, we introduce a grouping mechanism into MLA, enabling it to better adapt to sparse attention mechanisms.

We conduct a thorough evaluation of ASA using transformer models with 340M and 1.3B parameters, trained on 15B and 100B tokens sampled from the SlimPajama dataset. The trained models are evaluated across multiple task categories, including general common sense reasoning, long-context retrieval and long-context understanding. Experimental results show that ASA achieves performance on par with, or even exceeds, that of the full attention baseline, while outperforming existing sparse attention approaches. Moreover, compared to the full attention baseline, ASA reduces the KV-cache storage overhead by 50\%, offering a significant gain in memory efficiency without compromising model quality.


\section{Related Work}

With the increasing demand for processing long sequences in large language models, the quadratic complexity of vanilla attention has become a significant bottleneck. To address this challenge, a wide range of efficient attention mechanisms have been proposed~\cite{efficient-attention-sun2025}, where sparse attention is considered a promising approach.

Sparse attention mechanisms aim to enhance computational efficiency by selectively computing attention scores over a strategically chosen subset of key-value pairs, rather than exhaustively attending to all possible token interactions. By exploiting the intrinsic sparsity observed in attention distributions, these approaches substantially reduce computational complexity, often achieving sub-quadratic scaling, while preserving the model’s capacity to capture long-range contextual dependencies. Common sparse attention paradigms include attention sinks~\cite{stramingllm}, sliding window attention~\cite{swa-fu2025}, selective attention, and hybrid architectures that combine multiple sparsity patterns.

A salient feature of many sparsity-based methods is their training-free design, which allows seamless integration with pre-trained dense models without requiring additional fine-tuning. For instance, techniques such as Streaming LLM~\cite{stramingllm}, SnapKV~\cite{snapkv-li2024}, and PyramidKV~\cite{pyramidkv-cai2025} enforce a fixed-size cache during autoregressive decoding; as new key-value states are generated, they dynamically evict less salient states based on learned or heuristic attention scores. In contrast, methods like Quest~\cite{quest-tang2024}, infLLM~\cite{infllm-xiao2024}, ClusterKV~\cite{clusterkv-liu2025}, and RetroInfer~\cite{retroinfer-chen2025} retain the full sequence of key-value states and construct auxiliary indices to enable efficient retrieval. At each decoding step, only the most contextually relevant states are retrieved and attended to, thereby maintaining high fidelity while reducing computational overhead.

Further advances include approaches such as SeerAttention~\cite{seerattention-gao2025} and SeerAttention-R~\cite{seerattention-r-gao2025}, which introduce sparsity during the fine-tuning phase by incorporating distillation objectives that align dense softmax attention with sparse approximations. This regularization encourages the model to learn to prioritize semantically critical key-value pairs, thereby improving both efficiency and retrieval accuracy. Most recently, natively trainable sparse attention architectures, such as MoBA~\cite{moba-lu2025} and NSA~\cite{nsa-yuan2025}, have been proposed to explicitly optimize sparsity patterns during training, offering a more principled and adaptive exploration of attention sparsity beyond static or heuristic designs.

Despite their conceptual promise and demonstrated efficiency gains, existing sparse attention methods continue to face significant practical challenges that hinder their widespread deployment. Key-value evict strategies often exhibit degraded performance in long-context retrieval tasks~\cite{sparsefrontier-nawrot2025}. Conversely, retrieval-based approaches, while more robust to long-range dependencies, incur substantial computational overhead due to the latency associated with real-time indexing, sorting, and nearest-neighbor search operations; this overhead can outweigh their benefits in short-context or latency-sensitive scenarios. Moreover, many of these methods are not yet optimized for modern GPU architectures, whose memory hierarchies and parallel compute capabilities demand careful co-design to realize tangible acceleration.

\section{Background}

\vpara{Attention Mechanism.} The attention mechanism constitutes a fundamental component of contemporary language models. Given a query \( q_t \), it computes relevance scores with respect to all preceding keys \( K_t = [k_1, k_2, \dots, k_t] \), which are subsequently utilized to generate a weighted sum over the corresponding values \( V_t = [v_1, v_2, \dots, v_t] \). Formally, for an input sequence comprising \( t \) tokens, the attention mechanism with \(H\) query heads and group size \(G\) can be expressed as follows:
\begin{align*}
    o_t & = \text{Attn}(q_t, K_t, V_t) \\
        & = \Vert_{h=1}^{H}\{\text{Softmax}(\frac{q_t^{h} K_t^{\lfloor \frac{h}{G} \rfloor \top}}{\sqrt{d_k}})V_t^{\lfloor \frac{h}{G} \rfloor}\}
\end{align*}

\noindent where \(\Vert\) denotes the concatenation operation, \(q_t \in R^{H \times d_k}, K_t \in R^{t \times \lfloor \frac{H}{G} \rfloor \times d_k}, V_t \in R^{t \times \lfloor \frac{H}{G} \rfloor \times d_v}\), \(d_k\) and \(d_v\) represent the features dimension of keys and values respectively. It is evident that, within the attention mechanism, each query necessitates computation with all preceding key-value pairs. As the sequence length increases, the computational cost of attention progressively becomes the dominant factor in the overall model complexity, thereby presenting significant challenges for the efficient processing of long sequences.

\vpara{Native Sparse Attention.} To alleviate computational and memory access overhead in long-text scenarios, NSA introduces a native sparse attention mechanism. This mechanism decomposes the conventional attention operation into three distinct branches: sliding-window attention (\(o^{\text{swa}}\)), compressed attention (\(o^{\text{cmp}}\)), and selective attention (\(o^{\text{slc}}\)). Formally, NSA is defined as follows:

\begin{align*}
    o_t & = g^{\text{swa}}_t o^{\text{swa}}_t + g^{\text{cmp}}_t o^{\text{cmp}}_t + g^{\text{slc}}_t o^{\text{slc}}_t \\
    o^{\text{swa}}_t & = \text{Attn}(q_t, K_{t-s:t}, V_{t-s:t}) \\
    o^{\text{cmp}}_t & = \text{Attn}(q_t, \hat{K}_t, \hat{V}_{t}) \\
    o^{\text{slc}}_t & = \text{Attn}(q_t, K_{I_t}, V_{I_t}) \\    
\end{align*}

\noindent where \(g^{\text{swa}}\), \(g^{\text{cmp}}\) and \(g^{\text{slc}}\) are three gate scores to combine three attention branches. Specifically, for sliding-window attention, \(K_{t-s:t} = [k_{t-s}, \dots, k_t]\), where \(s\) denotes the sliding window size. For compressed attention, \(\hat{K}_t =[\text{cmp}(k_{1:B}),\dots,\text{cmp}(k_{mB-B+1:mB})]\), where \(\text{cmp}\) represents the compression operation, \(B\) is the compression block size, and \(m = \lfloor \frac{t}{B} \rfloor - 1\). For selective attention, \(K_{I_t} = \{K_{iB-B+1:iB}\}_{i \in I_t} \), where \(I_t = \text{Top-K}(\text{score}(q_t,\hat{K}_t))\) identifies the top \(K\) blocks based on the relevance scores between \(q_t\) and the compressed keys \(\hat{K}_t\). \(V_{t-s:t}\), \(\hat{V}_t\), and \(V_{I_t}\) can be obtained using the same method as \(K_{t-s:t}\), \(\hat{K}_t\), and \(K_{I_t}\).

\vpara{Multi-head Latent Attention.}  The Multi-head Latent Attention (MLA) mechanism was first introduced in DeepSeek-V2~\cite{deepseek-v2} and has demonstrated superior performance compared to conventional Multi-head Attention (MHA). MLA’s key innovation lies in its use of latent states and a reparameterization strategy that allows it to emulate MHA during training while effectively operating as Multi-query Attention (MQA) during inference. Formally, given an input \(x\), MLA first compresses \(x\) into a set of low-dimensional latent states \(c\), where \(\dim(c) \ll \dim(x)\). These latent states are then linearly projected to obtain key-value pairs: \(k = W_k c\) and \(v = W_v c\). Letting \(q = W_q x\), a simplified attention output at time step \(t\) can be expressed as  
\begin{align*}
    o_t = \bigl(\text{Softmax}(q_t k_{\le t}^\top) v_{\le t}\bigr) W_o.
\end{align*}
For MLA, this becomes  
\begin{align*}
    o_t &= \bigl(\text{Softmax}\bigl((x_t W_q)(c_{\le t} W_k)^\top\bigr) (c_{\le t} W_v)\bigr) W_o \\
        &= \bigl(\text{Softmax}\bigl((x_t W_q W_k^\top) c_{\le t}\bigr) c_{\le t}\bigr) (W_v W_o).
\end{align*}
By merging \(W_q\) and \(W_k\) into a single projection, and similarly combining \(W_v\) and \(W_o\), MLA requires storing only the compact latent states \(c\) during decoding, effectively reducing its memory footprint to that of MQA while retaining the expressive power of MHA during training.

\section{Rethinking Native Sparse Attention}

In this section, we provide a detailed analysis of the individual functions of these branches as well as their combinatorial effects within the NSA framework.

\begin{table*}[t]
\newcolumntype{?}{!{\vrule width 1pt}}
\newcolumntype{C}{>{\centering\arraybackslash}p{1.2cm}}
\centering
\renewcommand{\arraystretch}{1.0}
\resizebox{0.9\textwidth}{!}{
\begin{tabular}{l?c?ccccccc?c}
\toprule
\textbf{Model} 
& \makecell{\textbf{LAMB.}\\ppl \(\downarrow\)}
& \makecell{\textbf{LAMB.}\\acc \(\uparrow\)}
& \makecell{\textbf{PIQA}\\acc \(\uparrow\)}
& \makecell{\textbf{Hella.}\\acc\_n \(\uparrow\)}
& \makecell{\textbf{Wino.}\\acc \(\uparrow\)}
& \makecell{\textbf{ARC-e}\\acc \(\uparrow\)}
& \makecell{\textbf{ARC-c}\\acc\_n \(\uparrow\)}
& \makecell{\textbf{BoolQ}\\acc \(\uparrow\)}
& \makecell{\textbf{Avg.}} \\
\midrule
NSA & 44.34 & 31.03 & 64.31 & 34.78 & 51.07 & 44.07 & 23.29 & 58.06 & 43.80 \\
\midrule
\textit{Train from scratch}
& \multicolumn{1}{c?}{}
& \multicolumn{7}{c?}{}
& \multicolumn{1}{c}{} \\
\;- \textit{(w/o. slc)} & 40.58 & 31.24 & 63.66 & 34.70 & 52.33 & 43.98 & 22.70 & 55.81 & 43.49 \\
\;- \textit{(w/o. swa)} & 39.42 & 32.14 & 63.98 & 35.07 & 50.83 & 44.28 & 22.70 & 57.25 & 43.75 \\
\;- \textit{(w. alt)}  & 40.47 & 31.07 & 64.25 & 34.58 & 52.25 & 43.14 & 22.87 & 58.44 & 43.80 \\
\midrule
\textit{Traning free}
& \multicolumn{1}{c?}{}
& \multicolumn{7}{c?}{}
& \multicolumn{1}{c}{} \\
\;- \textit{(w/o. slc)} & 227.0 & 12.58 & 63.00 & 33.51 & 51.85 & 41.58 & 23.98 & 62.08 & 41.23 \\
\;- \textit{(w/o. swa)} & NAN & 0.0 & 51.85 & 25.90 & 49.96 & 25.97 & 26.79 & 38.26 & 31.25 \\
\midrule
\textit{NSA w/o. slc}
& \multicolumn{1}{c?}{}
& \multicolumn{7}{c?}{}
& \multicolumn{1}{c}{} \\
\;- \textit{(block size = 8)} & 40.23 & 31.71 & 63.93 & 35.27 & 52.80 & 43.01 & 23.46 & 57.58 & 43.97 \\
\;- \textit{(block size = 16)} & 40.58 & 31.24 & 63.66 & 34.70 & 52.33 & 43.98 & 22.70 & 55.81 & 43.49 \\
\bottomrule
\end{tabular}
}

\medskip

\resizebox{0.7\textwidth}{!}{
\begin{tabular}{l?ccc?ccc?ccc}
\toprule
\textbf{Model} 
& \multicolumn{3}{c?}{\makecell{\textbf{S-NIAH-1}\\(pass-key retrieval)}}
& \multicolumn{3}{c?}{\makecell{\textbf{S-NIAH-2}\\(number in haystack)}}
& \multicolumn{3}{c}{\makecell{\textbf{S-NIAH-3}\\(uuid in haystack)}} \\
\midrule
& \makecell{2k} & \makecell{4k} & \makecell{8k}
& \makecell{2k} & \makecell{4k} & \makecell{8k}
& \makecell{2k} & \makecell{4k} & \makecell{8k} \\
\midrule
NSA & 100.0 & 100.0 & 99.0 & 100.0 & 98.0 & 52.2 & 79.2 & 43.2 & 11.6 \\
\midrule
\textit{Train from scratch}
& \multicolumn{3}{c?}{}
& \multicolumn{3}{c?}{}
& \multicolumn{3}{c}{} \\
\;- \textit{(w/o. swa)} & 100.0 & 100.0 & 95.60 & 100.0 & 92.6 & 53.4 & 99.4 & 58.0 & 30.6 \\
\;- \textit{(w/o. slc)} & 27.6 & 12.6 & 6.4 & 30.2 & 17.2 & 8.0 & 31.6 & 14.4 & 7.4 \\
\;- \textit{(w. alt)}   & 100.0 & 100.0 & 100.0 & 100.5 & 100.0 & 97.8 & 83.4 & 55.2 & 22.0 \\
\midrule
\textit{Traning free}
& \multicolumn{3}{c?}{}
& \multicolumn{3}{c?}{}
& \multicolumn{3}{c}{} \\
\;- \textit{(w/o. swa)} & 0.0 & 0.0 & 0.0 & 0.0 & 0.0 & 0.0 & 0.0 & 0.0 & 0.0 \\
\;- \textit{(w/o. slc)} & 23.6 & 10.2 & 4.8 & 20.0 & 9.0 & 3.8 & 3.2 & 0.8 & 0.8 \\
\midrule
\textit{NSA w/o. slc}
& \multicolumn{3}{c?}{}
& \multicolumn{3}{c?}{}
& \multicolumn{3}{c}{} \\
\;- \textit{(block size = 8)} & 28.2 & 13.0 & 6.8 & 31.8 & 18.6 & 8.0 & 30.8 & 14.6 & 8.0  \\
\;- \textit{(block size = 16)} & 27.6 & 12.6 & 6.4 & 30.2 & 17.2 & 8.0 & 31.6 & 14.4 & 7.4 \\
\bottomrule

\end{tabular}
}

\caption{Ablation results of NSA on common-sense reasoning and in-context retrieval benchmarks. \textit{NSA (w/o. swa)}, \textit{NSA (w/o. slc)}, and \textit{NSA (w. alt)} denote removing sliding window attention, removing selective attention, and modifying NSA to alternately use sliding window attention and selective attention, respectively. \label{table: nsa ablation}}
\end{table*}

\subsection{Attention Modules Functional Analysis}

Empirical evaluations of NSA demonstrate that its sparse attention formulation consistently achieves lower language modeling losses compared to full attention baselines. This observation naturally invites deeper inquiry: What are the distinct functional roles of each of the three sparse attention branches within the NSA architecture? Moreover, which branch contributes most substantially to overall performance?

To systematically evaluate the functional contributions of individual sparse attention branches within the NSA framework, we train three 340M-parameter models on a 15B-token corpus: (1) the full NSA architecture, (2) NSA with the sliding window branch ablated, and (3) NSA with the selective attention branch ablated. In addition, we examine the impact of directly removing attention branches from a pre-trained NSA model on downstream performance: (4) removal of the sliding window branch from the pre-trained NSA, and (5) removal of the selective attention branch from the pre-trained NSA. Finally, to assess the role of compressed attention, we first remove the selective attention branch and then train the NSA model under varying block sizes: (6) NSA without selective attention using a block size of 8, and (7) NSA without selective attention using a block size of 16. The experimental results are presented in Table~\ref{table: nsa ablation}, and further details of the experimental setup are provided in Section~\ref{section: experiments}.

Based on these experiments, we draw the following conclusions:  
(a) Sliding window attention primarily impacts the model's performance on common-sense reasoning tasks. (b) Selective attention plays a crucial role in enhancing retrieval capabilities. (c) Compressed attention in NSA primarily functions as a supplementary mechanism to selective attention. (d) The concurrent use of sliding window and selective attention appears to diminish the retrieval capability of the selective attention branch.

By comparing experiments (1) and (4), we observe that removing window attention causes a significant decline in the model's performance metrics on common-sense reasoning tasks, corroborating conclusion (a). \footnote{Although experiment (2) indicates that removing window attention during pretraining has negligible impact on the model, further analysis reveals that some selective attention mechanisms degrade into window attention at pretraining stage.} Comparing experiments (1), (3), and (5) reveals that removing selective attention, either during pretraining or after training, leads to a significant drop in in-context retrieval task metrics, confirming conclusion (b).  Comparing experiments (1), (6), and (7) shows that even with finer-grained compressed attention, removing selective attention fails to effectively improve in-context retrieval performance, supporting conclusion (c). Finally, comparing experiments (1) and (2) reveals that removing window attention actually enhances the model’s in-context retrieval capability, confirming conclusion (d). We hypothesize that this occurs because the sliding window attention mechanism is more readily learned by the model, forming a shortcut that reduces reliance on selective attention during retrieval tasks, thereby weakening its effectiveness.

\subsection{Attention Modules Combination Analysis}

Excluding combinations of the three attention branches, the NSA framework applies a uniform sparsity rate to the selective attention branch at each layer, resulting in the retrieval of an identical number of key-value blocks. This design choice raises an important research question: Is an even distribution of sparsity across all layers optimal for selective attention?

Motivated by recent developments in hybrid attention architectures in open-source models, we investigated two alternative combination strategies:  (1) Applying the same sparsity level to the selective attention branches across all layers. (2) Alternating between compressed/selective attention and sliding window attention, i.e., removing selective attention from half of the layers and consolidating their computational workload into the remaining layers.

The experimental results are also shown in Table~\ref{table: nsa ablation}. It is evident that the alternating sliding window attention and selective attention (NSA with alt.) achieves superior contextual retrieval performance relative to the standard NSA architecture, while preserving comparable common-sense reasoning abilities. Additionally, this approach reduces the storage overhead of the KV-Cache by half.


\begin{table*}[t]
\newcolumntype{?}{!{\vrule width 1pt}}
\newcolumntype{C}{>{\centering\arraybackslash}p{1.2cm}}
\centering
\renewcommand{\arraystretch}{1.0}
\resizebox{0.8\textwidth}{!}{
\begin{tabular}{c?c?ccccccc?c}
\toprule
\textbf{Model} 
& \makecell{\textbf{LAMB.}\\ppl \(\downarrow\)}
& \makecell{\textbf{LAMB.}\\acc \(\uparrow\)}
& \makecell{\textbf{PIQA}\\acc \(\uparrow\)}
& \makecell{\textbf{Hella.}\\acc\_n \(\uparrow\)}
& \makecell{\textbf{Wino.}\\acc \(\uparrow\)}
& \makecell{\textbf{ARC-e}\\acc \(\uparrow\)}
& \makecell{\textbf{ARC-c}\\acc\_n \(\uparrow\)}
& \makecell{\textbf{BoolQ}\\acc \(\uparrow\)}
& \makecell{\textbf{Avg.}} \\
\midrule
\textit{340M params}
& \multicolumn{1}{c?}{}
& \multicolumn{7}{c?}{}
& \multicolumn{1}{c}{} \\
GQA & 36.40 & 34.33 & 62.95 & 34.65 & 50.99 & 43.73 & 23.55 & 52.51 & 43.24 \\
NSA & 44.34 & 31.03 & 64.31 & 34.78 & 51.07 & 44.07 & 23.29 & 58.06 & 43.80 \\
ASA & 40.47 & 31.07 & 64.25 & 34.86 & 52.25 & 44.28 & 23.29 & 58.44 & 44.06 \\
\midrule
\textit{1.3B params}
& \multicolumn{1}{c?}{}
& \multicolumn{7}{c?}{}
& \multicolumn{1}{c}{} \\
GQA & 14.99   &  47.56  & 69.31  & 49.59  &  54.54  &  55.30   &  26.96  &  56.91  & 51.45 \\
NSA & 12.29   &  50.44  & 71.06  & 51.67  &  55.56  &  57.07   &  26.71  &  58.20  & 52.96 \\
ASA & 11.21   &  51.73  & 71.33  & 51.73  &  55.01  &  56.52   &  27.13  &  58.26  & 53.10 \\

\bottomrule
\end{tabular}
}
\caption{Experiments results of GQA, NSA and ASA on common-sense reasoning benchmarks.\label{table: common sense}}
\end{table*}

\begin{table*}[t]
\newcolumntype{?}{!{\vrule width 1pt}}
\newcolumntype{C}{>{\centering\arraybackslash}p{1.2cm}}
\centering
\renewcommand{\arraystretch}{1.0}
\resizebox{0.65\textwidth}{!}{
\begin{tabular}{c?ccc?ccc?ccc}
\toprule
\textbf{Model} 
& \multicolumn{3}{c?}{\makecell{\textbf{S-NIAH-1}\\(pass-key retrieval)}}
& \multicolumn{3}{c?}{\makecell{\textbf{S-NIAH-2}\\(number in haystack)}}
& \multicolumn{3}{c}{\makecell{\textbf{S-NIAH-3}\\(uuid in haystack)}} \\
& \makecell{2k} & \makecell{4k} & \makecell{8k}
& \makecell{2k} & \makecell{4k} & \makecell{8k}
& \makecell{2k} & \makecell{4k} & \makecell{8k} \\
\midrule
\textit{340M params}
& \multicolumn{3}{c?}{}
& \multicolumn{3}{c?}{}
& \multicolumn{3}{c}{} \\
GQA & 100.0 & 100.0 & 100.0 & 100.0 & 83.2 & 54.6 & 97.2 & 90.8 & 33.0 \\
NSA & 100.0 & 100.0 & 99.0 & 100.0 & 98.0 & 52.2 & 79.2 & 43.2 & 11.6 \\
ASA & 100.0 & 100.0 & 100.0 & 100.0 & 100.0 & 99.8 & 83.4 & 62.6 & 52.6 \\
\midrule
\textit{1.3B params}
& \multicolumn{3}{c?}{}
& \multicolumn{3}{c?}{}
& \multicolumn{3}{c}{} \\
GQA & 100.0 & 100.0 & 100.0 & 100.0 & 100.0 & 100.0 & 84.2 & 93.0 & 64.4  \\
NSA & 100.0 &  99.8 & 98.8  & 100.0 & 99.8  & 66.0  & 89.6 & 78.8 & 65.0  \\
ASA & 100.0 & 100.0 & 100.0 & 100.0 & 100.0 & 100.0 & 87.4 & 79.4 & 62.0 \\
\bottomrule
\end{tabular}
}

\caption{Experiments results of GQA, NSA and ASA on in-context retrieval benchmarks.\label{table: in-context retrieval}}
\end{table*}

\begin{table*}[t]
\newcolumntype{?}{!{\vrule width 1pt}}
\newcolumntype{C}{>{\centering\arraybackslash}p{1.2cm}}
\centering
\renewcommand{\arraystretch}{1.0}
\resizebox{\textwidth}{!}{
\begin{tabular}{c ? c c c ? c c c ? c c c ? c c c ? c c ? c}
\toprule
\textbf{Model} & \multicolumn{3}{c?}{\textbf{Single-Doc QA}} & \multicolumn{3}{c?}{\textbf{Multi-Doc QA}} & \multicolumn{3}{c?}{\textbf{Summarization}} & \multicolumn{3}{c?}{\textbf{Few-shot}} & \multicolumn{2}{c?}{\textbf{Code}} & \textbf{Avg} \\
& NQA & QQA & MFQ & HQA & 2WM & Mus & GvR & QMS & MNs & TRC & TQA & SSM & LCC & RBP & \\
\midrule
\textit{340M params} & \multicolumn{3}{c?}{} & \multicolumn{3}{c?}{} & \multicolumn{3}{c?}{} & \multicolumn{3}{c?}{} & \multicolumn{2}{c?}{} &  \\
GQA           & 2.68 & 5.87 & 9.80 & 3.62 & 6.14 & 2.03 & 19.71 & 14.33 & 17.33 & 21.00 & 12.16 & 10.15 & 13.22 & 12.39 & 10.75\\
NSA           & 2.69 & 5.75 & 9.59 & 2.46 & 6.07 & 1.66 & 18.90 & 13.94 & 17.87 & 19.50 & 10.10 & 4.99 & 19.95 & 20.87 & 11.02 \\
ASA           & 2.82 & 6.25 & 10.64 & 3.95 & 6.20 & 2.45 & 16.47 & 14.23 & 18.01 & 33.00 & 12.96 & 8.43 & 21.72 & 20.19 & 12.67\\
\midrule
\textit{1.3B params} & \multicolumn{3}{c?}{} & \multicolumn{3}{c?}{} & \multicolumn{3}{c?}{} & \multicolumn{3}{c?}{} & \multicolumn{2}{c?}{} &  \\
GQA           & 2.92 & 7.77 & 13.52 & 5.53 & 8.80 & 3.13 & 21.65 & 15.22 & 20.33 & 37.50 & 32.34 & 18.68 & 22.72 & 20.81 & 16.49 \\
NSA           & 3.83 & 6.94 & 12.74 & 5.40 & 7.56 & 2.29 & 22.48 & 15.21 & 16.80 & 38.50 & 29.21 & 16.38 & 28.38 & 29.29 & 16.78 \\
ASA           & 3.39 & 8.35 & 14.29 & 4.46 & 8.09 & 2.96 & 23.51 & 15.89 & 18.71 & 54.00 & 26.87 & 16.47 & 27.49 & 30.96 & 18.25 \\
\bottomrule
\end{tabular}
}

\caption{Experiment results of GQA, NSA and ASA on long-context understanding benchmarks.\label{table: long-context understanding}}
\end{table*}

\section{Methodology}

In this section, we introduce our proposed method, Alternating Sparse Attention (ASA) and detail the algorithmic and engineering advancements incorporated in ASA over NSA, leading to improved model performance as well as enhanced training and inference efficiency.

\subsection{Alternating Sparse Attention}

In the preceding sections, our analysis of the various branches within NSA leads to the following conjectures: (1) Sliding-window attention plays a predominant role in minimizing language modeling loss. (2) The selective branch primarily facilitates long-context retrieval. (3) While the selective branch is essential, it is not required for all attention heads; employing computationally intensive selective branches for a subset of attention heads yields better performance than applying lightweight selective branches uniformly across all heads. 

Based on the conjecture above, we propose the Alternating Sparse Attention (ASA) mechanism, which introduces several architectural refinements to the baseline NSA framework, enhancing efficiency, scalability, and modeling capacity. Our improvements are structured as follows:

First, we redistribute the three attention branches originally integrated within each NSA layer across distinct layers of the model. This stratification ensures that each attention head specializes in a single sparsity pattern, thereby reducing interference and improving representational focus. Specifically, within each transformer layer, we sequentially apply the selective/compressed attention branch followed by the sliding window attention branch.

Second, we replace the Grouped Query Attention (GQA) mechanism, previously reliant on NSA, with Multi-head Latent Attention (MLA). Recognizing the critical role of sliding window attention in preserving local coherence and its measurable impact on language modeling loss. MLA enhances the expressiveness and contextual sensitivity of the sliding window branch by leveraging latent states, thereby improving modeling fidelity without compromising computational efficiency. 


Although Multi-Latent Attention (MLA) serves as a robust alternative to Grouped-Query Attention (GQA), it behaves identically to standard Multi-Head Attention (MHA) during training. Consequently, MLA remains incompatible with compression techniques and selective attention mechanisms that rely on shared key-value representations.  

To address this limitation, we introduce grouping within MLA specifically for the selective/compression branch—where all key-value pairs must be retained to support long-context retrieval. This modification yields Grouped-head Latent Attention (GLA), which replaces GQA in this branch. As a result, the retrieval capabilities of Adaptive Sparse Attention (ASA) are enhanced, enabling more effective processing of long-context sequences.

Formally, MLA can be expressed as:
\begin{align*}
    &\text{MLA}(q_{i,t}, c_{\le t}) =\\
    &\sum_{i=1}^{H} \text{Softmax}\big(q_{i,t} (c_{\le t} W_{i,k})^\top\big) W_{i,v} W_{i,o},
\end{align*}
where \(H\) denotes the number of attention heads, \(q_{i,t}\) represents the query at head \(i\) and time step \(t\), and \(c_{\le t}\) denotes the latent states up to time \(t\).  

During training, each attention head in MLA maintains independent query, key, and value projections, which prevents parameter sharing across heads and thus renders MLA incompatible with Native Sparse Attention (NSA).  

To adapt MLA for use with NSA, we further incorporate grouping into its architecture: multiple queries now share the same key and value projections. This leads to the following formulation of GLA:
\begin{align*}
    &\text{GLA}(q_{i,t}, c_{\le t}) = \\ 
    &\sum_{i=1}^{H/G} \sum_{j=1}^{G} \text{Softmax}\big(q_{iG+j,t} (c_{\le t} W_{j,k})^\top\big) W_{j,v} W_{iG+j,o},
\end{align*}
where \(G\) is the group size, and each group of \(G\) heads shares the same key and value projection matrices (\(W_{j,k}\) and \(W_{j,v}\)), while retaining distinct output projections (\(W_{iG+j,o}\)). In Appendix~\ref{sec: code}, we provide PyTorch-style pseudocode for ASA.

\subsection{Kernel Optimization}

The progress of sparse attention has been limited by the lack of hardware-efficient kernels, which makes it challenging to use sparse attention during the pre-training of models. To enable efficient pre-training with sparse attention, NSA modifies flash attention and introduces a kernel that is optimized for hardware efficiency in sparse attention scenarios. The main change in NSA is to move the partitioning strategy from the query sequence dimension to the head number dimension. However, this approach is restricted by the total number of attention heads, which prevents full utilization of computational resources.

Previous research shows that in sparse attention, the sets of key-value blocks retrieved by adjacent queries often have significant overlap. Based on this observation, we attempted to force consecutive queries to attend to the same key-value block during pre-training. Surprisingly, this adjustment had little effect on the model's performance. Building on this, we reintroduced sequential partitioning using NSA's attention kernel, with the added requirement that all queries in a block use the key-value block selected by the first query in that block. This change allows ASA to make better use of computational resources compared to NSA. The results of kernel optimization are shown in Appendix~\ref{sec: efficiency}.

\begin{table}[t]
\newcolumntype{?}{!{\vrule width 1pt}}
\newcolumntype{C}{>{\centering\arraybackslash}p{1.2cm}}
\centering
\renewcommand{\arraystretch}{0.8}
\resizebox{0.85\linewidth}{!}{
\begin{tabular}{l?cc?cc}
\toprule
                      & \multicolumn{2}{c?}{340M} & \multicolumn{2}{c}{1.3B} \\
                      & GQA/NSA       & ASA       & GQA/NSA       & ASA      \\ \midrule
\(n_{\text{layer}}\)  & \multicolumn{2}{c?}{21}   & \multicolumn{2}{c}{24}   \\
\(d_{\text{model}}\)  & \multicolumn{2}{c?}{1024} & \multicolumn{2}{c}{2048} \\
\(h_{\text{q}}\)      & \multicolumn{2}{c?}{16}   & \multicolumn{2}{c}{32}   \\
\(h_{\text{kv}}\)     & \multicolumn{2}{c?}{1}    & \multicolumn{2}{c}{2}    \\
\(d_{\text{vo}}\)      & \multicolumn{2}{c?}{128}  & \multicolumn{2}{c}{128} \\
\(d_{\text{ffn}}\)    & \multicolumn{2}{c?}{2816} & \multicolumn{2}{c}{5632} \\ \midrule
\(h_{\text{latent}}\) & -             & 256       & -             & 512      \\
\(d_{\text{qk}}\)      & 128           & 192       & 128           & 192      \\ \bottomrule
\end{tabular}
}
\caption{Parameter configuration for GQA, NSA, and ASA models with 340M/1.3B parameters.\label{table: config}}
\end{table}

\section{Experiments\label{section: experiments}}

Following the common practice in existing works, we evaluate ASA through common-sense reasoning tasks, in-context retrieval tasks, and long-context understanding tasks, comparing against group-query attention and native-sparse attention baseline.

\vpara{Setup} In our experiments, to ensure a fair comparison, all models were trained under identical conditions. We adopted the Llama architecture as the backbone and developed two model variants with 340M and 1.3B parameters, respectively. Detailed parameter configurations are shown in Table~\ref{table: config}. The models were trained on 15B and 100B tokens, which were sampled from the SlimPajama dataset. For optimization, we utilized the AdamW optimizer with a peak learning rate of $3 \times 10^{-4}$ and a minimum learning rate of $3 \times 10^{-5}$, a weight decay coefficient of 0.01, and a gradient clipping threshold of 1.0. The learning rate was maintained using a cosine schedule, with a warm-up period of 0.5B tokens, and a batch size of 0.5M tokens. All models utilized the Llama-2~\cite{Llama-2} tokenizer, which has a vocabulary size of 32,000. During training, the maximum context length was set to 8K tokens. To accelerate training, the key-value block retrieved from the first token is reused for every 4 consecutive tokens in ASA. For compression and selective attention, the block size is set to 16. Also, 64 blocks are selected per NSA layer, while 128 blocks are selected per pair of ASA layers. For comparison, we evaluate against the group query attention~\cite{gqa-ainslie2023} and native sparse attention~\cite{nsa-yuan2025}. 

\vpara{Common-Sense Reasoning}  
To assess the model's common-sense reasoning capabilities, we follow previous works~\cite{gdn-yang} and evaluate our model as well as baselines on several widely-used benchmarks. These include PIQA~\cite{piqa-bisk2019}, HellaSwag~\cite{hellaswag-zellers2019}, WinoGrande~\cite{winogrande-sakaguchi2019}, ARC-easy and ARC-challenge~\cite{ai2arc-clark2018}, SIQA~\cite{socialiqa-sap2019}, BoolQ~\cite{boolq-clark2019}, and LAMBADA~\cite{lambada-paperno2016}.

\vpara{In-Context Retrieval}  
For in-context retrieval tasks, we employ the Needle-In-A-Haystack Single (NIAH-S) benchmark from RULER~\cite{ruler-hsieh2024}, which comprises three tasks of increasing complexity: S-NIAH-1 (passkey retrieval), S-NIAH-2 (numerical needle in haystack), and S-NIAH-3 (word-based needle in haystack).

\vpara{Long Context Understanding}  
To evaluate long context understanding, we use 14 tasks from the LongBench~\cite{longbench-bai2024}. These tasks cover various aspects, including narrative comprehension~\cite{narrativeqa-kočiský2017} (Narrative QA), scientific understanding~\cite{qasperqa-dasigi2021} (QasperQA), multi-hop reasoning (MultiField QA, Hotpot QA~\cite{hotpotqa-yang2018}, 2WikiMulti QA~\cite{2wikimultiqa-ho2020}, Musique~\cite{musique-trivedi2022}), document summarization (GovReport~\cite{govreport-huang2021}, QMSum~\cite{qmsum-zhong2021}, MultiNews~\cite{multinews-fabbri2019}), as well as specialized tasks such as TRec~\cite{TREC-Li2002LearningQC}, Trivia QA~\cite{triviaqa-joshi2017}, SAMSum~\cite{samsum-gliwa-etal-2019}, LCC~\cite{LCC-mohler-etal-2016}, and RepoBench-P~\cite{repobench-liu2023}.

\vpara{Experiment Results} 
The experimental results of ASA and the baseline methods on common-sense reasoning, in-context retrieval, and long-context understanding benchmarks are presented in Tables~\ref{table: common sense},~\ref{table: in-context retrieval}, and~\ref{table: long-context understanding}. 

For common-sense reasoning tasks, ASA achieves slightly better performance than both GQA and NSA, highlighting the gains brought by replacing GQA with MLA. In context retrieval tasks, ASA clearly outperforms NSA, benefiting from the use of alternating hybrid window attention and selective attention. Additionally, with the integration of GLA, which increases the key-value dimensions during attention computation, ASA is able to surpass even the GQA baseline on the S-NIAH-2 task. On the S-NIAH-3 task, although ASA falls short of GQA at the 4k context length, it outperforms GQA at the 8k length. Finally, for long-context understanding tasks, ASA consistently outperforms both GQA and NSA across nearly all benchmarks.

\section{Conclusion}

In this work, we propose Alternating Sparse Attention (ASA), a novel sparse attention architecture that synergistically combines sliding-window and compress/selective attention mechanisms with multi-head and group-head latent attention enhancements. ASA achieves efficient long-context modeling for transformer-based large language models, matching or surpassing GQA and NSA in performance while reducing KV-cache storage by 50\%, thus providing a practical and scalable solution for memory-efficient language modeling.

\section{Limitation}

Although this work demonstrates that applying hybrid window attention and compressed selective attention in different layers yields better model performance than using them concurrently within the same layer, further exploration of optimal blending strategies remains necessary. For instance, it would be valuable to investigate whether there exists an optimal ratio and placement scheme for integrating hybrid window attention with compression and selective attention. Moreover, it also merits further study whether substituting sliding window attention with modern linear attention architectures could lead to models with greater expressive power.

\bibliography{custom}

\newpage

\appendix

\section{PyTorch-style pseudocode for ASA\label{sec: code}}

In Lsting~\ref{lst: ASA}, we present the pytorch-style pseudocode for ASA. In this code, the input to the attention module is denoted by \(x\). The matrices \(W_{c/q/k/v/g}\) represent learnable weights. Here, \(T\) refers to the number of tokens, \(HQ\) and \(H\) represents the number of attention heads for queries and key-values, \(d\) denotes the hidden dimension of the input, \(d_c\) represents the hidden dimension of the latents, \(d_p\) indicates the hidden dimension of the partial RoPE, and \(d_k\) and \(d_v\) correspond to the hidden dimensions of the key and value, respectively.

\lstset{
    language=Python,
    basicstyle=\ttfamily\footnotesize, 
    keywordstyle=\color{blue},
    stringstyle=\color{red},
    commentstyle=\color{green!70!black},
    showstringspaces=false,
    numbers=left,
    numberstyle=\tiny\color{gray},
    frame=tb,              
    framerule=1.5pt,       
    rulecolor=\color{black}, 
    breaklines=true
}

\begin{lstlisting}[language=Python,caption={PyTorch-style pseudocode for ASA},label={lst: ASA}]
def ASA_sliding_window_attention(
    x, W_q, W_c, W_p, W_k, W_v, W_g,
    T, HQ, d_c, d_p, d_k, d_v,
):
    q = (x @ W_q).view(T, HQ, d_k)
    q_nope, q_rope = split(q, [d_k - d_p, d_p], dim=-1)
    q_rope = RoPE(q_rope)
    q = cat([q_nope, q_rope], dim=-1)
    k_rope = RoPE((x @ W_p).view(T, 1, d_p))
    c = (x @ W_c).view(T, 1, d_c)
    k_nope = (c @ W_k).view(T, HQ, d_k - d_p)
    k = cat([k_nope, repeat(k_rope)], dim=-1)
    v = (c @ W_v).view(T, HQ, d_v)
    g = (c @ W_g).view(T, HQ)
    o = g * sliding_window_attention(q, k, v)
    return o

def ASA_cmpressed_selected_attention(
    x, W_q, W_c, W_p, W_k, W_v, W_g,
    T, B, HQ, H, d_c, d_p, d_k, d_v,
):
    q = (x @ W_q).view(T, HQ, d_k)
    q_nope, q_rope = split(q, [d_k - d_p, d_p], dim=-1)
    q_rope = RoPE(q_rope)
    q = cat([q_nope, q_rope], dim=-1)
    k_rope = RoPE((x @ W_p).view(T, 1, d_p))
    c = (x @ W_c).view(T, 1, d_c)
    k_nope = (c @ W_k).view(T, H, d_k - d_p)
    k = cat([k_nope, repeat(k_rope)], dim=-1)
    v = (c @ W_v).view(T, H, d_v)
    g_cmp, g_slc = split((c @ W_g).view(T, HQ * 2))
    # [T/B, H, d_k/d_v]
    k_cmp, v_cmp = compress(k, v, B) 
    I = topk(q, k_cmp)
    k_slc, v_slc = select(k, v, I)
    o = g_cmp * compressed_attention(q, k_cmp, v_cmp) \
        + g_slc * selected_attention(q, k_slc, v_slc)
    return o
\end{lstlisting}

\begin{table*}[h]
\newcolumntype{?}{!{\vrule width 1pt}}
\newcolumntype{C}{>{\centering\arraybackslash}p{1.2cm}}
\centering
\renewcommand{\arraystretch}{1.0}
\resizebox{\textwidth}{!}{
\begin{tabular}{c ? c c c ? c c c ? c c c ? c c c ? c c ? c}
\toprule
\textbf{Model} & \multicolumn{3}{c?}{\textbf{Single-Doc QA}} & \multicolumn{3}{c?}{\textbf{Multi-Doc QA}} & \multicolumn{3}{c?}{\textbf{Summarization}} & \multicolumn{3}{c?}{\textbf{Few-shot}} & \multicolumn{2}{c?}{\textbf{Code}} & \textbf{Avg} \\
& NQA & QQA & MFQ & HQA & 2WM & Mus & GvR & QMS & MNs & TRC & TQA & SSM & LCC & RBP & \\
\midrule
\textit{340M params} & \multicolumn{3}{c?}{} & \multicolumn{3}{c?}{} & \multicolumn{3}{c?}{} & \multicolumn{3}{c?}{} & \multicolumn{2}{c?}{} &  \\
NSA           & 2.69 & 5.75 & 9.59 & 2.46 & 6.07 & 1.66 & 18.90 & 13.94 & 17.87 & 19.50 & 10.10 & 4.99 & 19.95 & 20.87 & 11.02 \\
NSA (w. KO)   & 2.54 & 6.05 & 9.59 & 2.93 & 5.00 & 2.13 & 19.72 & 14.39 & 20.13 & 12.75 & 15.30 & 8.50 & 17.13 & 16.99 & 10.94\\
\bottomrule
\end{tabular}
}


\caption{Experiment results of NSA and NSA with kernel optimization (w. KO) on long-context understanding benchmarks.\label{table: kernel optimization}}
\end{table*}

\section{Kernel Optimization\label{sec: efficiency}}

\begin{figure}[t]
    \centering
    \includegraphics[width=\linewidth]{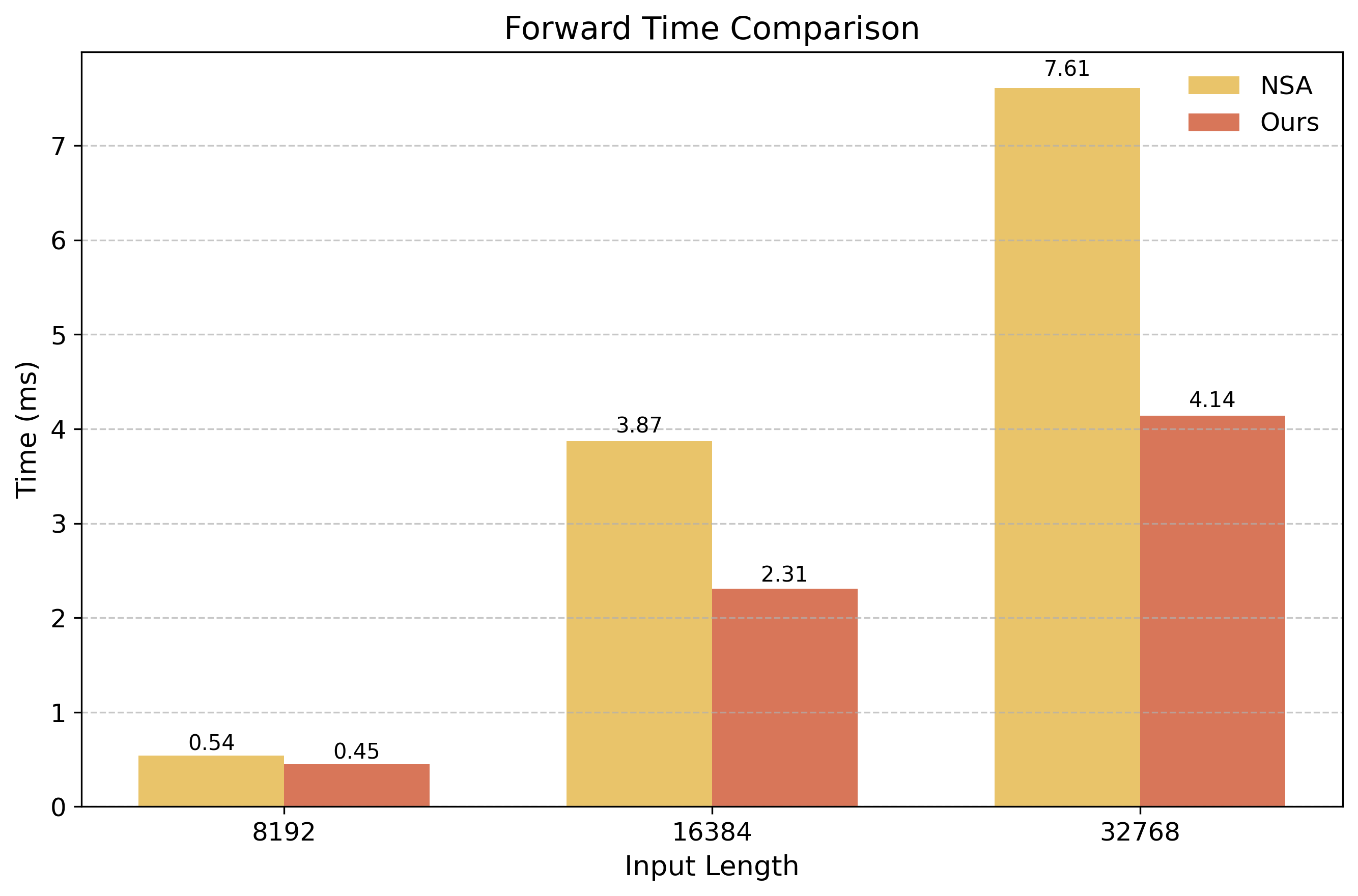}
    \caption{Computation time comparison for forward.\label{fig: forward}}
    
    \includegraphics[width=\linewidth]{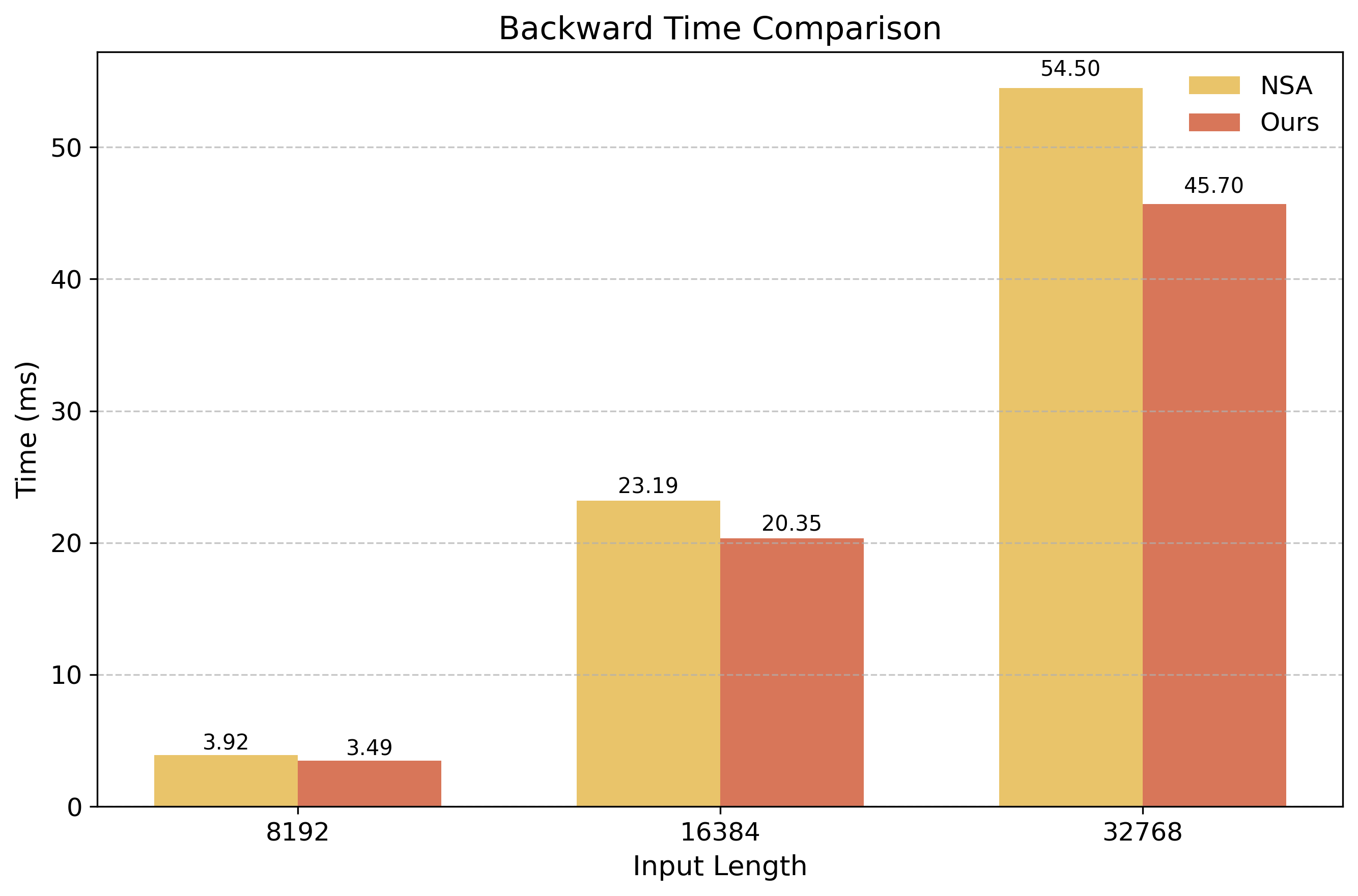}
    \caption{Computation time comparison for backward.\label{fig: backward}}
\end{figure}

We implemented our improved kernel based on the open-source NSA kernel\footnote{https://github.com/fla-org/native-sparse-attention} and compared it with the original version, in which consecutive four queries are guaranteed to select the same block. The forward and backward computation times were measured for sequence lengths of 8192, 16384, and 32768. All evaluations were conducted on a single H800 GPU. In the experiments, the batch size was set to 1, the number of kv groups was 1, and each kv group contained 16 query heads. The block size was set to 16, and each query selected 64 blocks (i.e., 1024 tokens). The experimental results are shown in Figures~\ref{fig: forward} and~\ref{fig: backward}. As can be seen, the optimized kernel reduces the forward computation time by approximately 30\% and the backward computation time by about 13\%. As shown in Table~\ref{table: kernel optimization}, we also evaluate NSA’s performance on the long-context understanding dataset under two settings: with and without kernel optimization. The results suggest that applying kernel optimization results in only a negligible reduction in model performance.

\end{document}